# Feature-Area Optimization: A Novel SAR Image Registration Method

Fuqiang Liu, Fukun Bi, Liang Chen, Hao Shi and Wei Liu

*Abstract*—This letter proposes a synthetic aperture radar (SAR) image registration method named Feature-Area Optimization (FAO). First, the traditional area-based optimization model is reconstructed and decomposed into three key but uncertain factors: initialization, slice set and regularization. Next, structural features are extracted by scale invariant feature transform (SIFT) in dual-resolution space (SIFT-DRS), a novel SIFT-Like method dedicated to FAO. Then, the three key factors are determined based on these features. Finally, solving the factor-determined optimization model can get the registration result. A series of experiments demonstrate that the proposed method can register multi-temporal SAR images accurately and efficiently.

*Index Terms*—SAR image registration, Area-based optimization model, initialization and regularization, scale invariant feature transform (SIFT), dual-resolution space

## I. Introduction

SYNTHETIC aperture radar (SAR) image registration is a fundamental process for various applications, such as change detection, image fusion, and multi-temporal analysis. However, it is more difficult to register SAR images than optical images as a result of multiplicative noise, which is referred to as speckle [1].

In general, feature-based registration methods are adopted to improve the adaptability of SAR image registration. Based on scale invariant feature transform (SIFT) [2], Schwind et al. [3] proposed the SIFT-Octave (SIFT-OCT) which improves the probability of correct matches by skipping the first octave of the scale space pyramid. You and Fu [4] proposed the bilateral filter SIFT (BF-SIFT) to fit SAR image registration by adopting an anisotropic Gaussian scale space. The above two methods can reduce the number of false matches, but they both need to decrease image resolution and thus lose information. Dellinger et al. [5] proposed SAR-SIFT, which adopted a new gradient definition, a new detector employing SAR-Harris scale-space and a new orientation assignment for SAR image registration. However, the results of SAR-SIFT still carry almost 10% false matches [5], which cause registration biases. These methods require an accurate feature extractor and a consistent feature descriptor to make sure that the same features in different pre-registration images can be matched accurately. Then, the geometric transformation can be executed based on a number of feature matches.

Unlike the feature-based methods, area-based methods [7] do not extract and match features. Therefore, they are not disturbed by false matches. These methods turn registration into an optimization problem and tune the geometric transformation based on the corresponding spatial relationship between images. If the optimization model can converge to a global extreme efficiently, it can be used to accurate SAR image registration. However, compared to feature-based methods, area-based methods are less commonly used in SAR image registration [6]. Under the conditions that the SAR images are usually large and occupied by serious speckles, these methods are computationally expensive and liable to get local optima [6].

To overcome the above problems, we propose a novel SAR image registration method named Feature-Area Optimization (FAO). In FAO, features are used to help area-based optimization model converge to a global optimum and reduce its complexity. First, we decompose the traditional area-based optimization model into three key but uncertain factors: initialization, slice set and regularization. Next, structural features are extracted by SIFT in dual-resolution space (SIFT-DRS). SIFT-DRS is a novel SIFT-like method dedicated to extracting features from pre-registration images fast. Then, the three factors are determined based on these features as follows: the initialization is set equal to the transformation matrix computed based on feature matches; the original images are replaced by slices of superposition; and the regularization item is determined based on the determined initialization. Finally, solving the factor-determined optimization model can get the SAR registration result.

In traditional area-based methods [7], the initialization is set empirically or close to zero, and the regularization cannot be determined if there is no prior information for SAR registration. The innovation of the proposed method is to use structural features extracted from pre-registration images to determine these factors. Thanks to the three key factors determined based on features, the area-based optimization model avoids converging to local optima. And by an optimizing process, the registration result of the proposed method can be better than the results of feature-based methods. FAO can be regarded as a composition of feature-based method and area-based method. A series of experiments show that the proposed method is more effective and more efficient for SAR image registration.

This work was supported by the National Natural Science Foundation of China (Grant No. 61171194)

Fu-qiang Liu, Liang Chen and Hao Shi are with School of Information and Electronic, Beijing Institute of Technology, Beijing 100081, China. (e-mail: fqliu@outlook.com; chenl@bit.edu.cn)
Fu-kun BI is with College of Information Engineering, North China University of Technology, Beijing 100144, China.

## II. OPTIMIZATION MODEL FOR SAR IMAGE REGISTRATION AND FACTOR DECOMPOSITION

In this section, we describe the affine transformation model and the area-based optimization model for SAR image registration. And then we propose the factor decomposition of the optimization model.

### A. Affine Transformation Model

The affine transformation model [5] is suitable for image registration. Its formula is shown as follows:

$$\begin{cases} x_2 = a_1 x_1 + b_1 y_1 + c_1 \\ y_2 = a_2 x_1 + b_2 y_1 + c_2 \end{cases} \quad (1)$$

where $(x_1, y_1)$ and $(x_2, y_2)$ denote the coordinates of the pre-registration images. (1) can be shaped into matrix form as follows:

$$P_2 = H \cdot P_1 \quad (2)$$

$$H = \begin{bmatrix} a_1 & b_1 & c_1 \\ a_2 & b_2 & c_2 \\ 0 & 0 & 1 \end{bmatrix} \quad (3)$$

with $P_i(x_i, y_i, 1)$ denoting the generic SAR pixel of image $I_i$ and $H$ denoting the transformation matrix. Though SAR is intrinsically two-dimensional, the third coordinates of $P_i$ is essential for obeying the rule of matrix multiplication and testing result. Generally, the purpose of SAR image registration is to estimate the coefficients of $H$.

### B. Area-Based Optimization Model for Registration

In traditional area-based methods [7], the Euclidean norm is adopted as the criterion to estimate the transformation matrix. Assuming $P$ denotes the generic SAR pixel, and $M_i = [P_1^i, P_2^i, ..., P_n^i]$ denotes the set of pixels in $I_i$, the optimization model is shown as follows [8]:

$$\min_H \frac{1}{m} \| I_2(H \cdot M_1) - I_1(M_1) \|^2 \quad (4)$$

with $m$ donating the number of the pixels involved in computation.

The extreme of (4) is the result of SAR image registration. However, as a result of the large scale of SAR images and speckles, solving (4) directly is computationally expensive and (4) is liable to converge to a local extreme.

### C. Factor Decomposition of Area-Based Optimization Model

Generally, if only parts of pre-registration images are involved in (4), solving the optimization model will be computationally cheaper than solving the model involving original SAR images. We name the collection of these parts as slice set.

Involving only parts of original images in computation is to estimate the global optimum by a similar local optimum. This operation causes a problem named over-fitting. Over-fitting problems can be solved by an appropriate regularization [10].

In addition, an accurate initialization for the optimization model can avoid the convergence problems and make solving the optimization model easily.

Based on the above states, we reconstruct (4) as (5) and decompose (5) into three factors: initialization, slice set and regularization.

$$\min_H \frac{1}{m} \sum_i \| D_2^i (H \cdot M_1^i) - D_1^i(M_1^i) \|^2 + \lambda \cdot reg \quad (5)$$

where $reg$ is the regularization item, $\lambda$ is weight decay parameter and $D$ denotes the slice of original image. The superscript $i$ is the index and the subscript shows which image the slice belongs to. All slices in (5) constitute a slice set $set$. The weight decay parameter $\lambda$ is considered with regularization together.

Define some operations of $D$: $D^i \cap D^j$ returns the acreage of the superposition; $sum(D^i, D^j)$ returns a new slice which involves full place of $D^i, D^j$; $area(D)$ returns the acreage of $D$.

We set two limits for $D$: one is that there is no superposition between any two slices of one certain image; another one is that some parts of the two slices with same index should be same. The first limit avoids repeat computation and the second one makes sure these slices can be registered. The above limits can be written as (6) (7), respectively.

$$D_{\text{image}}^i \cap D_{\text{image}}^j = 0, i \neq j \quad (6)$$

$$D_i^{index} \cap D_j^{index} \neq 0, i \neq j \quad (7)$$

Proportion denotes the ratio between slice set and pre-registration images. We use proportion to describe the property of the slice set. Define proportion as follows:

$$\text{proportion} = \frac{area\left(sum(D_1^1, \cdots D_1^n)\right) + area\left(sum(D_2^1, \cdots D_2^n)\right)}{area(I_1) + area(I_2)} \quad (8)$$

with $D_1^1, \cdots D_1^n, D_2^1, \cdots D_2^n \in set$. The proportion is the main property of slice set.

The three factors influence whether (5) can be solved efficiently. But the appropriate values or forms of the three factors are uncertain. To make (5) soluble, we should find an accurate initialization, determine an appropriate proportion of the slice set, and choose the form of the regularization item.

## III. FACTOR DETERMINATION BASED ON FEATURES EXTRACTED BY SIFT-DRS

This section proposes using structural features to determine initialization, slice set, and regularization. First, dual-resolution space (SIFT-DRS) is designed to extract the distinctive features from SAR images. Then, we detail how to determine the three factors based on these structural features.

### A. Feature Extraction by SIFT in Dual-Resolution Space

In this part, we propose an effective and efficient method, SIFT in dual-resolution space (SIFT-DRS), to extract distinctive features from pre-registration images. These features are devoted to determine the three key factors: initialization, slice set and regularization. The diagram of SIFT-DRS is shown in Fig. 1. There is no relationship between dual-resolution space and the Gaussian Pyramid, which means that we do not adapt the internal processes of SIFT.

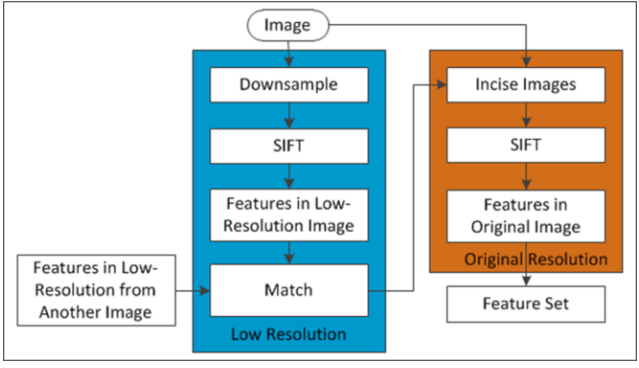

Fig. 1. Diagram of SIFT-DRS

SIFT-DRS is composed by two main operations, which are showed as "Low Resolution" and "Original Resolution" respectively in Fig.1. And this is the origin of dual-resolution space.

The image after dowmsampling is loosely regarded as low-resolution image. The image after downsampling should contain enough area for feature extraction. The minimum radius of the region where SIFT creates description is 11[2], and we need dozens of low-resolution features. As a result, we limit that the scale of the downsampled image is 128×128 at least. Defining $N$ as the downsampling rate and $L$ as the minimum scale of the original image, $N$ needs to satisfy the following:

$$\frac{L}{N} \geq 128 \quad (9)$$

In "Low Resolution", features are extracted by SIFT first, and then they are matched with the low-resolution features from another image. These matches demarcate the superposition of the two images roughly.

In "Original Resolution", the area where low-resolution feature matches demarcate is incised into some squares randomly. This step reduces the cost of further feature extraction. The size of each square is $\max\{N^2, 64 \times 64\}$ to make sure that there are enough area for feature extraction and description. Then features in original resolution are extracted from these squares by SIFT. The feature set is composed of these features.

By the dual resolution space, we can extract features from the pre-registration images more efficient than by SIFT in original images. And the feature set is used for determining the three key factors.

*B. Factor Determination*

In this part, we detail how to determine initialization, slice set and regularization based on the structural features.

First, matching pairs are found from the features extracted by SIFT-DRS. Then transformation matrix is computed by random sample consensus (RANSAC) [2].

As showed in Fig. 2, the RANSAC result, the transformation matrix, is set as the initialization. If two features from different image are matched, the neighborhoods of the features will satisfy (6) (7), and they can be chosen as the slices. We call the step of choosing which slices are involved in the slice set "Random Select and Proportion Limit". If the initialization is close to the global optimum, the registration result will be similar to the initialization. Based on the above statement, the form of regularization is determined by initialization.

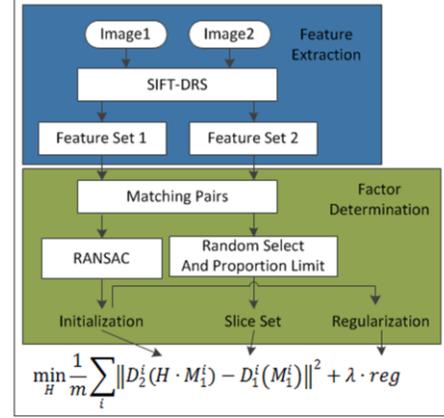

Fig. 2. Diagram of determining initialization, slice set and regularization based on structural features

*1) Initialization*

After SIFT-DRS, two feature sets of the pre-registration images are obtained. The matching pairs of these features are found first. Then those matching pairs are selected and the transformation matrix, $H_{trans}$, is computed by random sample consensus (RANSAC) [2]. Usually, feature-based methods regard the transformation matrix as the final registration result [2-6]. Here, we set $H_{trans}$ as the initialization, $H_0$.

*2) Slice Set*

The matching pairs mark part of the superposition of the two pre-registration images. In practice, the pre-registration images are cut into amounts of slices whose centers are the key points from those matching pairs. The sizes of the slices are set as 256×256 empirically. Each matching pair can determine two related slices, which is named related slice pair, from two images. Because the slices are chosen from the superposition, related slice pairs will satisfy (7). Then we select some of related slice pairs randomly to compose the slice set, only if slices from the same image satisfy (6). We use proportion, which is defined as (8), to limit how many slices we select.

*3) Regularization*

We use part of pre-registration images to replace the original images. In another word, the global optimum is estimated by an appropriate local optimum. This operation makes the optimization model over-fitting. As a result, a regularization item is added as (5). Previously, the regularization cannot be determined, because there is no prior information. If the initialization is close to the global optimum, the final result, registration matrix, will be similar to the initialization. Thus, we assume that $\|H - H_0\|$ is close to zero. Then, the above assumption can be regarded as prior information, and the regularization is set as $\|H - H_0\|^2$ in practice. The final form of (5) is showed as follows:

$$\min_H \frac{1}{m} \sum_i \left\| D_2^i(H \cdot M_1^i) - D_1^i(M_1^i) \right\|^2 + \lambda \|H - H_0\|^2 \quad (10)$$

where the weight decay parameter $\lambda$ controls the relative importance of the two items. We set $\lambda$ as 0.001. It may not be the optimal value for every SAR registration case, but it works well in our experiment.

Through the above three discussions, three key factors are determined. Solving (10) can get the registration matrix. All numbers determined in this section are valid for FAO. However, there are some uncertain parameters including downsampling rate $N$, proportion, and maximum generation. Maximum generation represents the maximum iteration times to solve (10). These parameters influence the effectiveness and efficiency of FAO.

## IV. EXPERIMENTS AND ANALYSIS

This section describes a series of experiments. How to choose the above uncertain parameters is discussed in experiment A. And experiment B validates our method by two registration examples and comparisons with previously works. All experiments are implemented by Matlab in a computer equipped with an Intel(R) Core(TM) i5-3470 and 4 GB of RAM. Moreover, the experimental data are acquired by the HJ-1C satellite with 5m ground resolution. A dataset containing 10 pairs of SAR images is carefully selected. These SAR images whose sizes are 4096×4096 contain various scenes, including mountain, sea, urban, river, forest and desert.

### A. Experiment on Parameter Determination

The parameters including downsampling rate $N$, proportion, and maximum generation are determined in experiment A.

*1) Downsampling Rate*

As mentioned previously, SIFT-DRS aims to extract reliable features rapidly. SIFT-DRS will be computationally cheap and the features will be similar to the results of SIFT, if $N$ is chosen appropriately. SIFT-DRSs in various $N$ are implemented throughout the dataset, and the mean times are recorded, which is shown in Fig. 3(a). The mean time how long operate SIFT throughout the dataset is 197 s.

To measure the accuracy of SIFT-DRS, (11) is employed to measure the similarity between features generated by SIFT-DRS and those extracted by SIFT.

$$\text{error} = \|C'_i - C_i\|^2 \quad (11)$$

where $C'_i$ is the coordinate of the feature of SIFT and $C_i$ is the coordinate of the related feature of SIFT-DRS. Fig. 3(b) shows the mean errors in different downsampling rate.

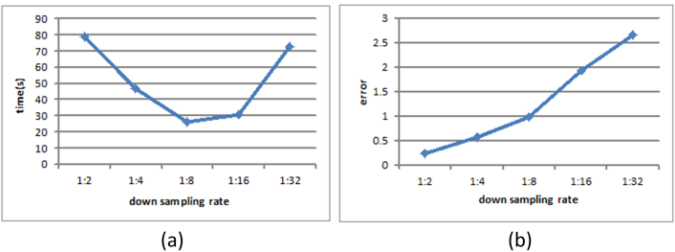

Fig. 3. Times and errors of SIFT-DRS in different downsampling rates

As Fig. 3 (a) shows, SIFT-DRS can drastically decrease the computational cost compared to SIFT. Nevertheless, a high downsampling rate does not mean a low cost because the squares in original resolution space are large. As Fig. 3(b) shows, the error is larger than 1 pixel when the downsampling rate is over 1:8. Based on this experiment, we set $N$ as 1:4 or 1:8 generally.

*2) Proportion*

This experiment is to find the appropriate proportion. Images from the above dataset are involved in this test. SIFT-DRS with downsampling rate 1:4 are completed first, and we only record the iteration step on solving (10). Root-mean-square error (RMSE) is adopted to measure the registration accuracy; it can be written as follows:

$$\text{RMSE} = \sqrt{\frac{\sum_{i=1}^{n}(x_i - x'_i)^2 + (y_i - y'_i)^2}{n}} \quad (12)$$

where $(x_i, y_i)$ and $(x'_i, y'_i)$ are the coordinates of the ith matching key points and n is the number of matching key points.

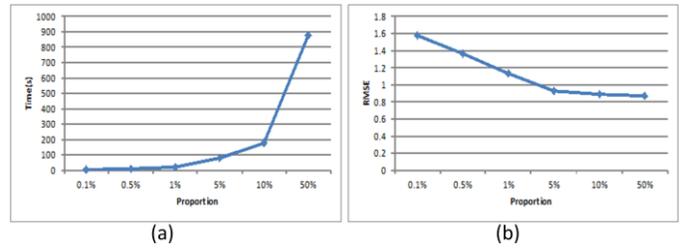

Fig. 4. (a) shows the relationship between time and proportion, and (b) shows the relationship between RMSE and proportion

The maximum generation is set as 1000. Then we record the computation times and the RMSEs in different proportions. Fig. 4(a) shows the relationship between time and proportion, and Fig. 4(b) shows the relationship between RMSE and proportion. As Fig. 4(b) shows, when the proportion is less than 5%, the RMSE is larger than 1. And a larger proportion leads to a smaller RMSE, but also leads to a more complex computation, which is shown in Fig. 4(a). We set the proportion as 5% so that we can get registration results with errors less than a pixel in a relatively short time.

*3) Maximum Generation*

In this experiment, downsamping rate is set as 1:4 and the proportion is set as 5%. Repeat the above experiment with different maximum generations, and record the times and the RMSEs. The result is shown in Table I.

TABLE I
TIME AND RMSE VARIES WITH GENERATION

| Generation | 1 | 10 | 50 | 100 | 200 | 300 |
|---|---|---|---|---|---|---|
| Time/s | - | 1 | 5 | 9 | 33 | 51 |
| RMSE | 1.87 | 1.55 | 1.26 | 1.22 | 0.895 | 0.893 |

As Table I shows, the RMSE converges with the generation. And when generation is over than 200, the RMSE varies barely. As a result, the maximum generation is chosen as 200. This experiment also shows that thanks to the accurate initialization, the optimization model converges quickly.

From experiment 1) to 3), the relatively optimal values of downsampling rate, proportion and maximum generation are determined. Downsampling rate influences the accuracy of the initialization and the time of feature extraction; proportion influences the accuracy of registration and the time of model solution; the maximum generation influence the time of model solution. These parameters can be set differently for different registration requirement.

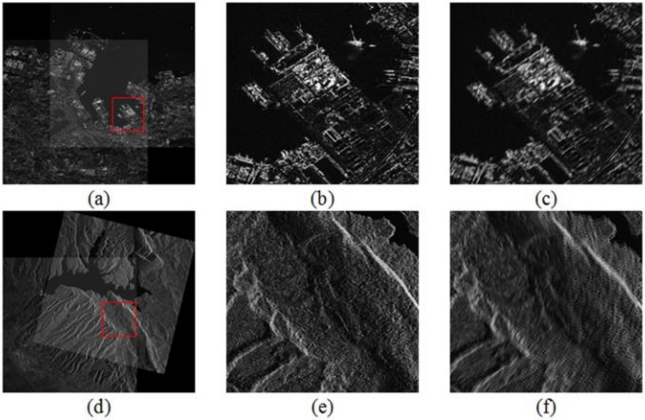

Fig. 5. Registration results of the proposed method. (b) is details of "red block place" in (a), and (e) is details of "red block place" in (d). (c) and (f) are results of cross correlation method

## B. Registration Result and Comparison with previous works

In this experiment, multi-temporal registration cases are carried out to validate our method. And we propose the result comparison with some previous works. Fig. 5(a) covers where the latitude is 35.446413 and the longitude is 139.694373, and Fig. 5(d) covers where the latitude is 42.661588 and the longitude is 41.781835. Here, downsampling rate is set as 1:4; proportion is set as 5%; and maximum generation is set as 200.

Fig. 5(a) and Fig. 5(d) are results of FAO. The superposition in co-registered images is marked by heightening the gray value factitiously. Fig. 5(b) and Fig. 5(e) are the details of "red block place" in Fig. 5(a) and Fig. 5(d), respectively. Fig. 5(c) and Fig. 5(f), which cover the same places of Fig. 5(b) and Fig. 5(e), are the results of cross correlation mothed.

TABLE II
ACCURACIES AND TIMES OF DIFFERENT METHODS FOR FIG. 5(A)

| Method | FAO | SIFT-OCT | BF-SIFT | Normalized Cross Correlation |
|---|---|---|---|---|
| RMSE | 1.01 | 1.77 | 1.37 | 5.78 |
| Time/s | 68 | 81 | 231 | 1751 |

TABLE III
ACCURACIES AND TIMES OF DIFFERENT METHODS FOR FIG. 5(B)

| Method | FAO | SIFT-OCT | BF-SIFT | Normalized Cross Correlation |
|---|---|---|---|---|
| RMSE | 0.83 | 1.31 | 1.16 | 4.69 |
| Time/s | 65 | 70 | 205 | 1622 |

These registration cases cannot be realized by solving (4) directly. But FAO can register these multi-temporal images efficiently and accurately. The same cases are realized by SIFT-OCT [3], BF-SIFT [4], and normalized cross correlation [10]. Table II shows RMSEs and times of different method for Fig. 5(a). And Table III shows RMSEs and times of different methods for Fig. 5(b). Table II and III show that the performance of FAO is better than the traditional methods, both in accuracy and efficiency.

FAO uses features to avoid convergence problem and reduce the computation time. Compared to traditional feature-based methods, FAO only extract features from downsampled images and parts of original images, which is the reason that FAO needs less time. And by solving area-based optimization model, the registration result is tuned to be better than results of traditional feature-based methods.

## V. CONCLUSION

In this letter, we introduce a novel SAR image registration method, FAO. FAO use features extracted by SIFT-DRS to determine initialization, slice set and regularization. From features to factors in area-based optimization model, FAO can be regarded as a composition of feature-based method and area-based method.

Thanks to feature-based method, area-based model avoids convergence problem and becomes computationally cheap. The result of a feature-based method (SIFT-DRS can be regarded a fast part of feature-based method) is tuned to be better by an optimizing process in area-based methods. And even though it is a composition of two type methods, it is more efficiently than traditional works.

There is still a problem for the proposed method. The optimal values of parameters in FAO are varied with the different registration application. In this letter, we set constant values and they work well in these registration cases. But a method on tuning the parameters automatically still deserves more works in the future.